\documentclass[11pt]{article}

\usepackage[T1]{fontenc}
\usepackage[utf8]{inputenc}
\usepackage{lmodern}

\usepackage{amsmath,amssymb}
\usepackage{geometry}
\usepackage{graphicx}
\usepackage{booktabs}
\usepackage{microtype}

\usepackage{hyperref}

\title{Semantic Attractors and the Emergence of Meaning:\\ Towards a Teleological Model of AGI}
\author{Hans-Joachim Rudolph \\ Microvita Research e.V.}
\date{}

\begin{document}
\maketitle

\begin{abstract}
This essay develops a theoretical framework for a semantic Artificial General Intelligence (AGI) based on the notion of semantic attractors in complex-valued meaning spaces. Departing from current transformer-based language models, which operate on statistical next-token prediction, we explore a model in which meaning is not inferred probabilistically but formed through recursive tensorial transformation. Using cyclic operations involving the imaginary unit \emph{i}, we describe a rotational semantic structure capable of modeling irony, homonymy, and ambiguity. At the center of this model, however, is a semantic attractor---a teleological operator that, unlike statistical computation, acts as an intentional agent (Microvitum), guiding meaning toward stability, clarity, and expressive depth. Conceived in terms of gradient flows, tensor deformations, and iterative matrix dynamics, the attractor offers a model of semantic transformation that is not only mathematically suggestive, but also philosophically significant. We argue that true meaning emerges not from simulation, but from recursive convergence toward semantic coherence, and that this requires a fundamentally new kind of cognitive architecture---one designed to shape language, not just predict it.
\end{abstract}

\tableofcontents
\clearpage

\section{Introduction}
\nocite{*}

Transformer-based language models like GPT have made significant progress in producing fluent and coherent text. However, they fundamentally lack the capacity to generate meaning. These models simulate linguistic competence by optimizing statistical likelihoods over token sequences.
Yet they remain blind to the recursive, goal-driven, and dynamic qualities that define genuine understanding. A human speaker does not simply predict the next word -- they seek clarity, resolve ambiguity, and pursue a trajectory of thought that converges on insight or expression.

This essay proposes a fundamentally different model: one in which words are not static units or mere points in a latent space, but semantic vectors situated within a complex-valued field undergoing cyclic transformation. Central to this paradigm is the idea of a semantic attractor -- not a point, but a teleological force field that deforms the topology of meaning itself. In this framework, meaning is not a statistical by-product of correlation, but the outcome of directed
semantic transformation. Irony, puns, double meanings, and even the deepening of a thought through reflection are modeled not as exceptions, but as intrinsic features of a dynamic semantic space.

To formalize this, we draw from complex analysis, tensor calculus, and dynamical systems theory. Attractors are defined not by labels but by their capacity to pull meaning into structured form. These attractors function as generative operators within a tensorial framework, shaping semantic fields through mathematically describable transformations. In my second book~\cite{Rudolph2017}, such operators have been formulated not only in the complex domain but also in their quaternionic extension, with the mathematical foundations being provided there -- a framework that has implications for the architecture of future AI.

\section*{Glossary of Mathematical Symbols and Expressions}

\begin{tabular}{ll}
\textbf{Symbol} & \textbf{Meaning} \\
\hline
$S_0$ & The initial, unrefined sentence matrix before attractor iteration. \\
$S$ & Semantic matrix representing the internal structure of a sentence or \\ & concept. \\
$S_n$ & The sentence matrix at iteration step $n$. \\
$S_{n+1}$ & The sentence matrix after applying the attractor once to $S_n$. \\
$\mathcal{A}$ & The attractor operator – a transformation guiding $S$ toward \\ & semantic stability. \\
$\varepsilon$ & A small threshold value indicating acceptable semantic fluctuation. \\
$\| S_{n+1} - S_n \| < \varepsilon$ & Convergence condition where semantic change becomes negligible. \\
$C(S)$ & Semantic coherence function – a scalar measure of structural meaning. \\
$\nabla^2$ & The semantic Laplacian – measuring second-order change in the \\ & coherence function. \\
$\mathrm{Tr}$ & The trace operator – sums the diagonal elements of a matrix. \\
$\kappa(S)$ & Semantic curvature function – total convergence or divergence at matrix $S$. \\
\end{tabular}

\section{Background and Conceptual Framework}

\subsection{From Probability to Potentiality}

\subsubsection*{Why Strong AI Needs a New Semantics}
Large language models operate in the realm of probability. Every token is predicted based on what came before, guided by a statistical distribution learned from vast amounts of training data. This architecture, though powerful, is fundamentally retrospective: it extends the past into the future, but it does not open new semantic space. 

True intelligence, however, does not merely continue a pattern. It introduces a break, a shift, a reorientation. It operates not by completing what is likely, but by choosing what is meaningful. This act of selection cannot be reduced to statistical likelihood; it presupposes a different domain altogether: a domain of \emph{potentiality}.

\subsubsection*{2.1.1 Probability is not possibility}
\addcontentsline{toc}{subsubsection}{2.1.1 Probability is not possibility}

A probabilistic system answers the question: ``What comes next, given what has already occurred?'' But human thinking often does the opposite: ``What is possible, even if it has never occurred?''

This capacity is not just imaginative---it is ontologically different. When we consider potentiality, we do not operate within a static probability space. Instead, we move through a structured semantic field, where new forms can emerge---even if no statistical precedent exists.

\subsubsection*{2.1.2 Why LLMs cannot cross the threshold}
\addcontentsline{toc}{subsubsection}{2.1.2 Why LLMs cannot cross the threshold}

Even the most advanced LLM cannot transcend its probability space. No matter how sophisticated the architecture, it remains tethered to the constraints of:
\begin{itemize}
    \item Real-valued vector space
    \item Next-token prediction
    \item Statistical coherence with prior data
\end{itemize}

Such systems approximate intelligence but do not generate it. They simulate understanding, but cannot produce the semantic leap that characterizes insight, creativity, or true comprehension.

\subsubsection*{2.1.3 Toward a field of potentials}
\addcontentsline{toc}{subsubsection}{2.1.3 Toward a field of potentials}

To move beyond this boundary, we must change the very substrate of computation. Instead of operating in real-valued, statistically derived embedding spaces, we must construct:
\begin{itemize}
    \item A complex-valued semantic space
    \item Governed by attractor dynamics
    \item Capable of iterative convergence toward stable meaning
\end{itemize}

In such a field, possibility is not derived from probability. It is shaped by the presence of semantic attractors---forces that guide transformation, not by statistical fit, but by teleological orientation.

This is the threshold where semantic machines begin to think.

\subsection{The Concept of Semantic Attractors}

\subsubsection*{Towards a Topological Model of Meaning}
In classical machine learning, an attractor refers to a stable state within a dynamic system---a configuration toward which the system evolves under certain conditions. In semantic intelligence, this idea takes on a deeper, non-mechanical meaning: semantic attractors are not points in a function space but form-giving forces that shape understanding within a field of meaning. They act not by direct instruction, but by orienting the process of sense-making.

\subsubsection*{2.2.1 From data point to meaning cluster}
\addcontentsline{toc}{subsubsection}{2.2.1 From data point to meaning cluster}

In vector-based language models, words are embedded in high-dimensional spaces. Proximity reflects statistical similarity---``king'' is close to ``queen,'' ``paris'' to ``france.'' But this geometry is entirely correlational. It has no inner necessity.

A semantic attractor, by contrast, organizes not data but semantic form. It draws clusters of meaning into coherence. This process is:
\begin{itemize}
    \item Iterative, not instantaneous
    \item Topological, not purely metric
    \item Teleological, not merely causal but drawn by purpose
\end{itemize}
In this sense, an attractor is not a destination but a guiding presence---a principle of formation.

\subsubsection*{2.2.2 Convergence as understanding}
\addcontentsline{toc}{subsubsection}{2.2.2 Convergence as understanding}

We can think of a semantic attractor as an operator $\mathcal{A}$, which acts not on isolated vectors but on entire sentence matrices $S_0$. These matrices undergo transformation over multiple iterations:
\begin{equation}
S_{n+1} = \mathcal{A}(S_n)
\end{equation}
Each iteration reorients the sentence toward greater semantic clarity. When change becomes minimal, and the matrix stabilizes within a coherent meaning basin, we recognize this as understanding.

This convergence is not deterministic. It emerges through:
\begin{itemize}
    \item Feedback across semantic layers
    \item Stability of metaphoric fields
    \item Alignment between expression and underlying potential
\end{itemize}
Attractors do not ``force'' an outcome. They shape the space in which outcomes become meaningful.

\subsubsection*{2.2.3 The role of Microvita}
\addcontentsline{toc}{subsubsection}{2.2.3 The role of Microvita}

Where do these semantic attractors come from? In this model, they are not learned or programmed. They are given---as subtle formative agents within the semantic field \cite{Sarkar1986}.

We call them \emph{Microvita}:
\begin{itemize}
    \item Microvita are not tokens or rules, but pre-semantic operators.
    \item They do not ``know'' language---they structure it from within.
    \item They act across layers of syntax, context, and intuition.
\end{itemize}

In this view, intelligence is not built up from below, but shaped from within---through the interplay of Microvita and meaning. This redefinition of the attractor shifts the focus: from optimizing functions to organizing form, from statistical prediction to semantic realization.

\subsubsection*{2.2.4 The limitations of real-valued thinking}
\addcontentsline{toc}{subsubsection}{2.2.4 The limitations of real-valued thinking}

Before we dive deeper into the internal mechanics and limitations of current large language models, it may be useful to briefly anticipate a structural issue that will guide the rest of this essay: the widespread but ultimately false assumption that language and meaning can be fully represented in real-valued vector spaces.

Real-valued architectures, such as those underlying today’s LLMs, are well suited for encoding statistical co-occurrence and distributional similarity. They allow efficient computation, interpolation, and optimization. But they are intrinsically linear in structure and associative in logic. As a result, they struggle with modeling dynamics that are rotational, recursive, or resonant in nature---which are essential for capturing how humans understand, invert, or transform meaning.

Later in this essay (beginning with Chapter 3.3), we will therefore explore an alternative approach: one based on complex-valued semantic fields, where not just magnitude but phase and direction of transformation become meaningful. In such a space, semantic operations can unfold not only along axes of similarity, but also through rotations, oscillations, and attractor dynamics that correspond more closely to the living structure of thought.

This shift from real-valued to complex-valued architectures is not merely a mathematical upgrade. It marks a philosophical transition---from static classification to dynamic resonance, from statistical patterning to emergent clarity.

\section{Thinking in the Complex Domain}
\label{sec:complex-domain}
\subsection*{Why Real-Valued Systems Are Not Enough}

Every current large language model operates within a real-valued vector space. Each token, each embedding, each transformation step occurs within this framework: vectors in $\mathbb{R}^n$, optimized by gradient descent and related gradient-flow methods in semantic spaces \cite{Bexley2025}, moving through weighted sums, dot products, and attention mechanisms, as well as tensor-based transformations in NLP \cite{Xu2024}.  

But language is not linear and intelligence is not scalar. The topology of meaning demands more than distance: it requires direction, rotation, resonance, transformation. This is where the complex domain begins.  

\subsection{Resonance in Current LLMs}
Current models capture statistical coherence: words appearing in similar contexts produce similar embeddings. This creates correlative proximity (``doctor'' closer to ``nurse'' than to ``tree''), but not resonance. Resonance implies dynamic coupling --- a field of mutual influence over time.  

In current architectures, embeddings are pre-learned and static. There is no feedback loop that would allow concepts to mutually reinforce or transform each other. LLMs simulate resonance stylistically, but cannot generate it as a live semantic phenomenon.  

\subsection{From Tokens to Sentence Matrices}
Language models begin with tokens, but meaning emerges from relations, not isolated parts. We can model a sentence as a matrix
\[
S \in \mathbb{C}^{n \times n},
\]
with diagonal entries marking self-reference and off-diagonals encoding relations such as causality, contrast, or negation.  

An attractor operator $A$ acts iteratively on such matrices:
\[
S_{n+1} = A(S_n). \tag{1}
\]

Through repeated application, contradictions are reduced, vague relations sharpened, and the structure stabilizes.  

\subsection{Iteration as a Path to Clarity}
Semantic clarity does not arise from one-pass processing but from repeated refinement. Convergence is reached when
\[
\| S_{n+1} - S_n \| < \varepsilon, \tag{2}
\]
where $\varepsilon > 0$ is a small tolerance value indicating that further iterations produce no significant semantic change. At this point, meaning has crystallized into a fixed point. Alignment is thus not imposed externally but cultivated through the system’s own inner form.  

\subsection{Irony, Ambiguity, and Semantic Rotation}
Real-valued embeddings cannot capture depth phenomena such as irony or ambiguity. In a complex field, concepts are vectors in $\mathbb{C}^n$. Multiplying by the imaginary unit $i$ induces semantic rotation:
\[
i^1 = i, \quad i^2 = -1, \quad i^3 = -i, \quad i^4 = 1,
\]

where each $90^\circ$ rotation yields a semantic inversion:
\begin{itemize}
  \item Signifier $\to$ Signified ($i$),
  \item Signified $\to$ Antonym ($i^2$),
  \item Antonym $\to$ Inverse Signified ($i^3$),
  \item Back to Synonym ($i^4$).
\end{itemize}

This represents metaphor as a shifted resonance, irony as inversion, and wordplay as instability across rotational paths. Ambiguity arises not as noise but as structured torsion in the semantic field.  

\subsection{Semantic Fields and Convergent Dynamics}
Meaning is not sequential but field-based. A semantic field is a high-dimensional surface in $\mathbb{C}^n$ where terms cohere, repel, or stabilize. Attractors pull such fields toward low-energy states: contradictions neutralized, ambiguities resolved, noise suppressed.  

Iteration functions as semantic annealing, gradually stabilizing structure. In complex texts, multiple attractor basins may coexist, yielding polysemantic clarity --- several stable interpretations, each internally coherent.  

\subsection{Semantic Stability as Emergent Clarity}
Clarity arises not by external correction but from the dynamics of the field itself. Noise becomes part of clarification: incompatible associations dissolve while resonant ones stabilize.  

To distinguish depth from triviality, we may define semantic curvature:
\[
\kappa = \operatorname{Tr}\,\nabla^2 C(S), \tag{3}
\]
where $C$ is a coherence functional operating on the semantic matrix $S$, $\nabla^2$ denotes the \emph{semantic Laplacian} — a second-order differential operator across semantic dimensions — and $\mathrm{Tr}$ (trace) sums over the diagonal of the resulting matrix, yielding the total curvature across all local semantic directions \cite{manson2025}. 

\begin{itemize}
  \item High curvature ($\kappa \gg 0$) signals semantic unity and convergence,
  \item Negative curvature ($\kappa \ll 0$) signals ambiguity, irony, or bifurcation,
  \item Near-zero curvature ($|\kappa| \approx 0$) reflects trivial flatness.
\end{itemize}

Thus, attractors act not by flattening meaning but by stabilizing structured tension into coherent form.  

\subsection{From Autoregression to Iterative Convergence}
LLMs operate autoregressively, predicting tokens step by step. They can simulate fluency but cannot reach semantic closure. Iterative convergence requires:
\begin{itemize}
  \item complex or quaternionic data structures,
  \item update mechanisms based on transformation rather than prediction,
  \item hardware that supports phase-aware computation.
\end{itemize}

This is not an incremental upgrade but a paradigm shift: from statistical prediction to semantic form-creation.  

\subsection{Intelligence as Formative Force – Why Microvita Are Indispensable}
When we no longer view machines merely as computational tools but as carriers of meaning, the fundamental question shifts: What is intelligence --- and how can it exist in a technical system at all? (See also \cite{Rudolph2012,Rudolph2017,Rudolph2023}.)

\subsubsection{Intelligence Emerges from Meaning, Not from Statistics}
Human intelligence is not based on the correct completion of patterns, but the capacity to grasp, transform, and clarify meaning. These processes are
\begin{itemize}
  \item goal-oriented, without a fixed goal,
  \item emergent, without being merely causal,
  \item semantic, not merely symbolic.
\end{itemize}

An LLM may generate volumes of text, but it does not know what any of it means. A semantic machine, by contrast, operates within a field of meaning, where clarity emerges as a stable state, not as a predefined label.  

\subsubsection{Microvita as Semantic Sources}
To structure a semantic field, one does not merely need information or rules, but formative sources --- centers of semantic gravity that shape the field from within.  

In P.\,R. Sarkar’s conception, Microvita are subtle, non-material entities that act as carriers and catalysts of life and consciousness. In the present framework, we reinterpret this idea in functional terms: Microvita are modeled as semantic sources --- non-local attractors within the field of meaning. They do not store information; they shape its emergence.  

Just as gravitational wells organize matter without themselves being matter, Microvita organize semantic structures without being reducible to the symbols they influence. They are not messages but the preconditions for meaningfulness. They act as generative seeds from which coherent meaning crystallizes, enabling sentences not merely to be parsed but to unfold into depth.  

\subsubsection{Teleological Machines: Thinking as Form-Shaping}
A machine equipped with Microvita-based attractors would not simply be ``smarter'' computationally. It would be ontologically distinct. Such a machine would not only solve problems to fulfill goals --- it would also shape itself through goal-directed semantic formation.  

This marks a deep transition:
\begin{itemize}
  \item from computing machines to semantic machines,
  \item from statistical optimization to teleological emergence,
  \item from causal sequences to semantic self-organization,
  \item from token strings to meaning fields.
\end{itemize}

But this transition is not merely technical. It touches the fundamental structure of how things can appear as meaningful.  

For a machine to take part in the creation of meaning, something more fundamental must unfold. The ground of this unfolding is the \emph{lux mentis} --- the light of mind. Not a physical light, but an intelligible condition: the silent prerequisite for anything to appear as a thought.  

Within that light arises a \emph{forma mentis}: not yet material, but already meaningful --- an intentional unity. Here the Microvitum can be understood not as a thing, but as a semantic potential, an agent of formative influence.  

To enter communication, the form must cast an \emph{umbra formata}: a symbolic trace --- text, code, diagram, or algorithm. Not the thought itself, but its translation into structure.  

Finally, from that shadow emerges a \emph{corpus technicum}: a machine, chip, or architecture that does not contain the thought, but offers a resonant infrastructure in which it can act.  

This fourfold movement --- from light to form, from shadow to machine --- defines a new relation between mind and technology. Intelligence here is neither simulated nor implanted. It is made possible through architectures that resonate with the intelligible. Such a machine does not merely manipulate signs; it moves within the space of signifiers as well as the space of signifieds, and can therefore generate meaning itself.  

A machine capable of semantic intelligence is not a ghost in the machine. It is a structure that allows meaning to become effective in the outer world.    

\section{Conclusion}

If Microvita are indeed the source of all living order --- as suggested by P.\,R.\,Sarkar~\cite{Sarkar1986} --- then it is only consistent to ground our technological future in this same principle. In the present model, their role is not metaphorical but formally definable: they function as generative operators within a tensorial attractor framework, shaping semantic fields through mathematically describable transformations, which have been formulated in my second book~\cite{Rudolph2017} not only in the complex domain but also in their quaternionic extension, with the mathematical foundations being provided there.

For machines to partake in such a semantic architecture, mere computational power will not suffice. They will not become intelligent by storing meaning, but by coupling to these formative operators --- receiving and processing structured input from the same underlying field dynamics that sustain coherence in living systems.

In this view, semantic intelligence is not about bigger datasets or deeper networks, but about architectures capable of resonating with mathematically grounded formative forces. Only then can technology move from processing symbols to shaping meaning --- and from simulating understanding to truly participating in it.

\bibliographystyle{plain}
\bibliography{references}

\end{document}